\documentclass[sigconf,nonacm,natbib=false]{acmart}

\usepackage[T1]{fontenc}
\usepackage[utf8]{inputenc}
\usepackage{newtxtext,newtxmath}
\usepackage{graphicx}
\usepackage{booktabs}
\usepackage{tabularx}
\usepackage{array}
\usepackage{amsmath}
\usepackage{mathtools}
\usepackage{microtype}
\usepackage{float}
\setcopyright{none}
\renewcommand\footnotetextcopyrightpermission[1]{}
\acmConference{}{}{}
\acmYear{}
\copyrightyear{}
\acmDOI{}
\acmISBN{}

\usepackage[
  backend=biber,
  style=numeric,
  sorting=none,
  giveninits=true,
  maxbibnames=3,
  minbibnames=3,
  doi=false,
  isbn=false,
  url=false
]{biblatex}
\DeclareNameAlias{default}{family-given}
\DeclareFieldFormat{title}{#1}
\DeclareFieldFormat[article,inproceedings,misc,thesis]{title}{#1}
\DeclareFieldFormat{journaltitle}{#1}
\DeclareFieldFormat{booktitle}{#1}
\DeclareFieldFormat{pages}{#1}
\DeclareFieldFormat{volume}{#1}
\DeclareFieldFormat{number}{#1}

\newbibmacro*{afpgnn:authors:title}{%
  \printnames{author}%
  \setunit{\addperiod\space}%
  \printfield{title}%
  \setunit{\addperiod\space}}

\DeclareBibliographyDriver{article}{%
  \usebibmacro{bibindex}%
  \usebibmacro{begentry}%
  \usebibmacro{afpgnn:authors:title}%
  \printfield{journaltitle}%
  \setunit{\addcomma\space}%
  \printfield{year}%
  \iffieldundef{volume}
    {}
    {\setunit{\addcomma\space}\printfield{volume}%
     \iffieldundef{number}{}{\printtext{\mkbibparens{\printfield{number}}}}}%
  \iffieldundef{pages}
    {}
    {\setunit{\addcolon\space}\printfield{pages}}%
  \finentry}

\DeclareBibliographyDriver{inproceedings}{%
  \usebibmacro{bibindex}%
  \usebibmacro{begentry}%
  \usebibmacro{afpgnn:authors:title}%
  \printfield{booktitle}%
  \setunit{\addcomma\space}%
  \printfield{year}%
  \iffieldundef{pages}
    {}
    {\setunit{\addcolon\space}\printfield{pages}}%
  \finentry}

\DeclareBibliographyDriver{misc}{%
  \usebibmacro{bibindex}%
  \usebibmacro{begentry}%
  \usebibmacro{afpgnn:authors:title}%
  \printfield{howpublished}%
  \setunit{\addcomma\space}%
  \printfield{year}%
  \finentry}

\DeclareBibliographyDriver{thesis}{%
  \usebibmacro{bibindex}%
  \usebibmacro{begentry}%
  \usebibmacro{afpgnn:authors:title}%
  \printfield{type}%
  \setunit{\addcomma\space}%
  \printlist{institution}%
  \setunit{\addcomma\space}%
  \printfield{year}%
  \finentry}

\renewbibmacro*{in:}{}

\newcolumntype{Y}{>{\centering\arraybackslash}X}
\newcolumntype{L}{>{\raggedright\arraybackslash}X}

\title{Semantic Representation Learning of Scientific Literature Based on Adaptive Feature and Graph Neural Network}

\author{Hongrui Gao}
\affiliation{%
  \institution{Beijing Key Laboratory of Intelligent Communication Software and Multimedia, School of Computer Science (National Pilot Software Engineering School), Beijing University of Posts and Telecommunications}
  \city{Beijing}
  \postcode{100876}
  \country{China}}

\author{Yawen Li}
\authornote{Corresponding author.}
\email{warmly0716@126.com}
\affiliation{%
  \institution{School of Economics and Management, Beijing University of Posts and Telecommunications}
  \city{Beijing}
  \postcode{100876}
  \country{China}}

\author{Meiyu Liang}
\affiliation{%
  \institution{Beijing Key Laboratory of Intelligent Communication Software and Multimedia, School of Computer Science (National Pilot Software Engineering School), Beijing University of Posts and Telecommunications}
  \city{Beijing}
  \postcode{100876}
  \country{China}}

\author{Zeli Guan}
\affiliation{%
  \institution{Beijing Key Laboratory of Intelligent Communication Software and Multimedia, School of Computer Science (National Pilot Software Engineering School), Beijing University of Posts and Telecommunications}
  \city{Beijing}
  \postcode{100876}
  \country{China}}

\author{Zhe Xue}
\affiliation{%
  \institution{Beijing Key Laboratory of Intelligent Communication Software and Multimedia, School of Computer Science (National Pilot Software Engineering School), Beijing University of Posts and Telecommunications}
  \city{Beijing}
  \postcode{100876}
  \country{China}}

\begin{abstract}
Because most scientific literature data are unlabeled, semantic representation learning based on unsupervised graphs has become crucial. To enrich scientific-literature features, this paper proposes a semantic representation learning method based on adaptive features and graph neural networks. By introducing adaptive feature processing, scientific-literature features are considered globally and locally. The graph attention mechanism weights and aggregates features of scientific documents connected by citation relations, so that correlations among different documents can be expressed more effectively. In addition, an unsupervised graph neural network semantic representation learning method is proposed. By comparing the mutual information between positive and negative local semantic representations of scientific literature and the global graph semantic representation in the latent space, the graph neural network captures local and global information and improves semantic representation learning. Experimental results show that the proposed method is competitive for scientific literature classification.
\end{abstract}


\thanks{This work was supported by the National Natural Science Foundation of China (62172056, U22B2038, 62192784).}

\begin{document}
\maketitle

\section{Introduction}

Scientific literature resources are increasingly abundant on the Internet. The amount of data is large, redundancy is limited, and many records cannot be discarded arbitrarily. Scientific literature contains valuable information, but existing indicators and categories are often too broad to cover and extract all useful semantics. Representation learning is connected to recommendation and intelligent decision tasks, where compact and robust features improve downstream utility \cite{zhou2022filterenhanced}. Scholar clustering studies show that academic objects may have dynamic interests and evolving relationships, which motivates richer representations for literature data \cite{li2023multiviewscholar}. Self-supervised recommendation, dataset distillation, and interpretable decision models further show the value of representation learning in downstream applications \cite{xia2021selfsupervised,zhang2025td3,li2019interpretable}. Data-driven analysis has also been used to study innovation and entrepreneurial identity in broader decision contexts \cite{zhou2020creative}. Therefore, how to extract valuable information from scientific and technological documents and build efficient semantic representations is a primary problem for classification, search, and recommendation.

Traditional methods mainly rely on expert experience to construct artificial features. In scientific and technological literature search, the title, abstract, and body of a document can be used to build a reverse index \cite{zhao2021abstract}. However, when new text data are added, rebuilding the index is complex. In document search systems, bag-of-words models \cite{huang2020shorttext}, topic models \cite{wang2020lda}, and vector models have been used to construct document vectors. When the vocabulary of the bag-of-words model increases, each sentence can contain only a small number of terms, which leads to sparse matrices and seriously affects memory and computing resources \cite{li2014lpv,li2017distributedekf,li2016tobitkalman}. Mainstream methods map data into vector spaces and operate on vectors to complete specific tasks \cite{deng2019vectorspace,liang2020abstractive}. Cross-media retrieval work also highlights the need to align scientific and technological information across heterogeneous media while preserving semantic consistency \cite{li2022semanticsadversarial}. However, many document-processing methods based on vector models depend on word-frequency information, so document similarity depends heavily on common words and cannot fully distinguish semantic ambiguity.

When learning semantic representations of scientific and technological documents, features are often fused into a single channel. This makes semantic features insufficiently rich, weakens the ability to capture influential text features, introduces redundant information, and allows irrelevant information to affect semantic representation learning. Adaptive text feature processing can alleviate the limitations of single-source word vectors \cite{fan2020adaptive}. Graph-oriented approaches, including heterogeneous graph attention, modularity-based community detection, and teacher-student graph learning for incomplete features and structures, also show that structural and semantic information can complement one another \cite{hu2019heterogeneous,yang2016modularity,huo2023t2gnn,jin2022heterogeneous}.

Semantic representation based on deep learning has recently received widespread attention \cite{li2017variance,ren2022attention}. Neural network models can automatically learn semantic features from scientific-document corpora \cite{xu2013imagefusion,li2017distributedconsensus} and express them as dense vectors, thereby supporting search tasks. However, neural language models usually focus on textual semantics \cite{meng2016consensus,jin2019similarity} and may ignore relationships among documents. Some researchers have addressed incomplete information through multi-agent and consensus mechanisms \cite{peng2009average,meng2013tracking,meng2015robust}. More studies integrate different features to support deep learning tasks \cite{li2013region,kou2018hashtag,li2022fuel}.

This paper proposes an unsupervised graph neural network method \cite{wei2019boosting,tan2022overview} to extract citation relationships from scientific-document graphs and fuse them with textual semantic features. Most existing studies use supervised graph neural networks to learn representations \cite{zhang2021gcntext,li2021heterogeneouslatent,shen2021gcnattention}. Cross-graph node classification indicates that graph learning is also important when data are distributed across multiple graphs or institutions \cite{guan2021federatedgraph}. For specific tasks, supervised methods require many labeled data, which increases training cost and strengthens the coupling between learned representations and downstream tasks \cite{khosla2019comparative}. In contrast, unsupervised graph neural networks can directly learn text features from unlabeled scientific-literature networks and transfer them to downstream tasks, including classification, search, and recommendation.

This paper proposes an unsupervised semantic representation learning method for scientific documents based on adaptive feature processing and graph neural networks (AFPGNN). The main contributions are as follows.

\begin{enumerate}
  \item A semantic representation learning method based on adaptive features and graph neural networks is proposed. The adaptive feature method considers global and local document information to enrich scientific-document features.
  \item To improve representation learning, graph attention is introduced to model relationships among scientific-document features. The semantic representation of a document is related to adjacent nodes and does not require receiving global information from all documents.
  \item An unsupervised graph neural network semantic representation learning method is proposed for unlabeled and scalable large-scale graphs. By comparing mutual information between positive and negative local representations and the global graph representation in latent space, the method improves semantic representation learning.
\end{enumerate}

\section{Related Work}

In natural language processing, semantic representations of scientific and technological documents can be obtained through pre-training models, including Word2Vec and FastText \cite{thavareesan2020sentiment}, ELMo-style contextual word embeddings \cite{popa2019implicit}, BERT \cite{devlin2018bert}, and SciBERT \cite{beltagy2019scibert}. Retrieval-oriented language-model pre-training further improves dense retrieval and document matching \cite{xiao2022retromae}. Supervised cross-modal retrieval also shows that federated learning can improve representation alignment under distributed data constraints \cite{li2024supervisedcrossmodal}. Stochastic quantization further addresses communication efficiency in federated learning settings \cite{li2022stochasticquantization}. In semantic representation learning based on deep learning, a key problem is how to construct a suitable neural network structure. Common methods include convolutional neural networks, attention mechanisms, recurrent neural networks, and pre-training with TextCNN \cite{yuan2020financial,hu2018gatherexcite,wang2019convolutional}. Attention mechanisms can dynamically focus on important input fragments. For example, attention-based explanation methods generate directional correlation scores from attention weights \cite{liu2021attention}.

By learning scientific-document features through deep neural networks, semantic representation learning can obtain strong data representation ability and learn general prior knowledge independently of specific tasks \cite{minae2021deep,li2022fuel}. Introducing semantic representation learning into scientific and technological document processing can therefore uncover latent value in document data.

Scientific-literature networks are high-dimensional. How to express high-dimensional document networks as low-dimensional semantic representation vectors is a central problem in graph embedding \cite{wu2020comprehensive,wu2021comprehensive}. Factorization, random walks, and deep learning are three major graph-embedding directions. Matrix-factorization approaches build matrices that measure pairwise similarity in different ways \cite{yu2019constrained}. Random-walk sentence embeddings model word generation probability with angular distance between word and sentence embeddings \cite{ethayarajh2018unsupervised}. Hypergraph-based scientific-publication representation further suggests that high-order semantic similarity can strengthen publication modeling \cite{li2026semanticsimilarity}. Convolutional neural networks have also been used for microblog topic content search \cite{nan2019microblog}.

Because most scientific literature data are unlabeled, unsupervised graph learning is important. Existing unsupervised graph learning often depends on random-walk objectives and is sensitive to parameter selection \cite{sato2020expressive}. Encoders can impose inductive bias so that adjacent nodes have similar representations. Federated self-adaptive information-network representation indicates that graph representations can also be learned under distributed and privacy-aware settings \cite{li2026fedsin}. Training encoders to distinguish representations and statistical relationships of interest is an important strategy for unsupervised semantic learning. Contrastive methods are central to many word-embedding and text-code representation models \cite{neelakantan2022textcode}. By maximizing mutual information between local document representations and global graph representations, graph neural networks can capture both local and global information.

\section{AFPGNN Based on Adaptive Feature and Graph Neural Network}

\begin{figure*}[t]
  \centering
  \includegraphics[width=\textwidth]{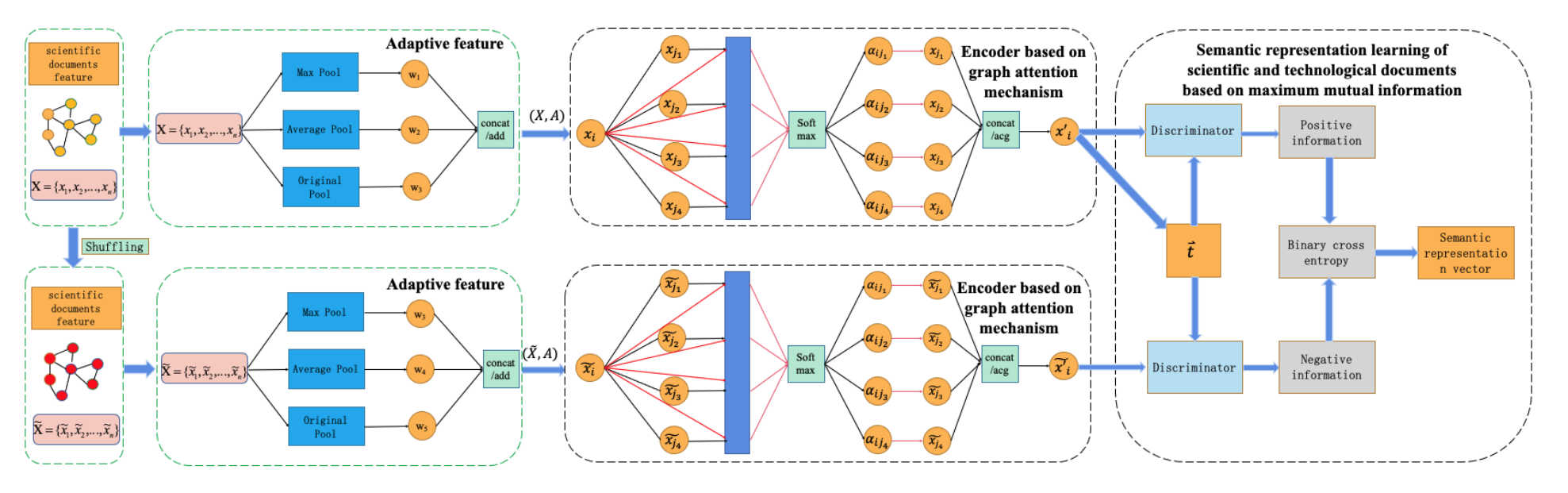}
  \Description{Framework diagram of AFPGNN with adaptive features, graph attention encoder, and mutual-information learning.}
  \caption{Learning method of semantic representation of scientific and technological documents based on adaptive feature and graph neural network.}
  \label{fig:framework}
\end{figure*}

To solve the problem that scientific-literature data are unlabeled, an unsupervised semantic representation learning method of scientific literature based on adaptive features and graph neural networks (AFPGNN) is proposed. Adaptive features are used to consider both global and local textual information, enriching the feature information of scientific documents. A graph attention encoder learns node representations, and unsupervised graph neural network representation learning solves unlabeled and scalable large-graph problems. The framework of the AFPGNN method is shown in Figure~\ref{fig:framework}.

\subsection{Adaptive Feature}

In the adaptive feature part, feature weights are added so that the method can process scientific-document features with specific weights according to their distribution. The adaptive feature processing part does not need to learn parameters. The feature processing formula is
\begin{equation}
F = \alpha_1 F_{\mathrm{original}} + \alpha_2 F_{\max} + \alpha_3 F_{\mathrm{avg}},
\end{equation}
where the normalized weight is
\begin{equation}
\alpha_i = \frac{e^{w_i}}{\sum_j e^{w_j}},
\quad i = 1,2,3;\; j = 1,2,3.
\end{equation}
Here $\alpha_i$ is the normalized weight and $w_i$ is the characteristic weight of scientific and technological documents.

Maximum pooling $F_{\max}$ is used to extract local features of scientific and technological documents:
\begin{equation}
F_{\max} = \max(X).
\end{equation}
Average pooling $F_{\mathrm{avg}}$ is used to extract global characteristics:
\begin{equation}
F_{\mathrm{avg}} = \mathrm{avg}(X).
\end{equation}
Here $\max(\cdot)$ finds the maximum eigenvalue of scientific documents and $\mathrm{avg}(\cdot)$ finds the average eigenvalue. $F_{\mathrm{original}}$ retains the original characteristics. The maximum-pooling part focuses on local information, while the average-pooling part focuses on overall information.

\subsection{Encoder Based on Graph Attention Mechanism}

A new feature is generated in the encoder based on the graph attention mechanism. Suppose the node feature dimension is $F'$, and this feature can be expressed as
\[
X' = \{x'_1, x'_2,\ldots,x'_N\}, \quad x'_i \in \mathbb{R}^{F'}.
\]
In the graph attention layer, a weight matrix acts on scientific-document nodes, and the attention coefficient is calculated using the self-attention mechanism:
\begin{equation}
e_{ij} = a(Wx_i, Wx_j).
\end{equation}
Here $e_{ij}$ is the impact of the characteristics of document $j$ on scientific document $i$. Softmax is introduced to the neighbor node $j$ of document node $i$:
\begin{equation}
\alpha_{ij} = \mathrm{softmax}_j(e_{ij})
= \frac{\exp(e_{ij})}{\sum_{m\in N_i}\exp(e_{im})}.
\end{equation}
After obtaining the weight matrix between connection layers, the output layer of the feedforward neural network is processed using the LeakyReLU function. Combining Equations~(5) and~(6), the complete attention mechanism is
\begin{equation}
\alpha_{ij} =
\frac{\exp\left(\mathrm{LeakyReLU}\left(a^T[Wx_i \Vert Wx_j]\right)\right)}
{\sum_{m\in N_i}\exp\left(\mathrm{LeakyReLU}\left(a^T[Wx_i \Vert Wx_m]\right)\right)}.
\end{equation}
The regularized attention coefficient between document nodes can then be used to predict each node feature:
\begin{equation}
x'_i = \mathrm{PReLU}\left(\sum_{j\in N_i}\alpha_{ij}Wx_j\right).
\end{equation}
Here $W$ is the feature weight matrix, $\alpha_{ij}$ is the attention joint coefficient, and $j\in N_i$ represents the neighbor nodes of $i$. The PReLU activation function adaptively learns rectifier parameters and improves accuracy at negligible additional cost.

To stabilize self-attention learning, multiple attention heads are used. Equation~(8) is transformed by introducing $K$ independent attention mechanisms:
\begin{equation}
x'_i = \mathrm{PReLU}\left(\frac{1}{K}\sum_{k=1}^{K}\sum_{j\in N_i}\alpha^k_{ij}W^kx_j\right).
\end{equation}
Here $K$ represents the number of attention heads, $\alpha^k_{ij}$ represents the $k$th attention mechanism, and $W^k$ represents the feature weight matrix under the $k$th attention mechanism.

\subsection{Semantic Representation Learning Based on Maximum Mutual Information}

The semantic representation learning method based on maximum mutual information uses an unsupervised graph contrastive learning strategy to address unlabeled and scalable large-scale graphs. By comparing mutual information between positive and negative local document node representations and the global graph representation in the latent space, the graph neural network captures local and global information.

The attention layer generates node embedding $x'_i$, and the global information centered on a node is not just the node itself. To obtain the graph-level summary vector $t$, a readout function $F:\mathbb{R}^{N\times F}\rightarrow \mathbb{R}^{F}$ aggregates node features:
\begin{equation}
t = F(E(X,A)).
\end{equation}
The graph-level representation of scientific documents is obtained by
\begin{equation}
F(X) = \frac{1}{N}\sum_{i=1}^{N}x'_i.
\end{equation}
A discriminator $D:\mathbb{R}^{F}\times\mathbb{R}^{F}\rightarrow\mathbb{R}$ is used to maximize local mutual information. $D(x_i,t)$ is the probability score assigned to the summary information:
\begin{equation}
D(x_i,t)=\sigma(x_i^T Wt).
\end{equation}
The random function $S:\mathbb{R}^{N\times F}\times \mathbb{R}^{N\times N}\rightarrow \mathbb{R}^{M\times F}\times \mathbb{R}^{M\times M}$ is used to generate negative samples. This process keeps the adjacency matrix of citation relationships unchanged and shuffles document-node features. The marginal distribution and joint distribution of local and global features have a Jensen-Shannon (JS) divergence positively correlated with mutual information. Binary cross entropy is used as the loss:
\begin{equation}
\begin{aligned}
\mathcal{L} =
\frac{1}{N+M}\bigg(
&\sum_{i=1}^{N}\mathbb{E}_{(X,A)}
\left[\log D(x_i,t)\right] \\
&+
\sum_{j=1}^{M}\mathbb{E}_{(\widetilde{X},A)}
\left[\log\left(1-D(\widetilde{x}_j,t)\right)\right]
\bigg).
\end{aligned}
\end{equation}

\section{Experimental Results and Analysis}

\subsection{Dataset}

AFPGNN is used for semantic representation learning of scientific and technological documents on the Cora and Citeseer data sets. Scientific-document classification is used as the downstream task. Accuracy, Macro-F1, and recall are used to evaluate performance. In the data sets, nodes represent scientific documents and edges represent citation relationships between documents. Table~\ref{tab:dataset} shows information about the Cora and Citeseer data sets.

\begin{table}[t]
  \caption{Cora and Citeseer dataset information.}
  \label{tab:dataset}
  \centering
  \small
  \begin{tabularx}{\linewidth}{@{}lYY@{}}
    \toprule
    & Cora & Citeseer \\
    \midrule
    Node & 2708 & 3327 \\
    Edge & 5429 & 4732 \\
    Feature & 1433 & 3703 \\
    Class & 7 & 6 \\
    Training & 140 & 120 \\
    Validation & 500 & 500 \\
    Test & 1000 & 1000 \\
    \bottomrule
  \end{tabularx}
\end{table}

\subsection{Comparison Method}

To illustrate effectiveness, AFPGNN is compared with the following methods.
\begin{itemize}
  \item DeepWalk randomly selects a node to jump with a certain probability and learns node vector representations through convergence stability.
  \item MLP uses input, output, and hidden layers.
  \item GCN obtains feature information from neighboring nodes, averages node characteristics, and inputs them into the neural network.
  \item GAE uses convolution kernels and a graph decoder in the encoder process.
  \item VGAE replaces the encoder process with GCN convolution on the basis of GAE and replaces the decoder with a graph decoder.
  \item GraphSAGE samples between nodes and aggregates information through multi-layer aggregation functions.
  \item MoNet uses learnable kernel functions to improve semantic representation learning.
\end{itemize}

\subsection{Experiment 1: Comparative Experiment on Cora and Citeseer}

Accuracy, Macro-F1, and recall are used to evaluate AFPGNN in scientific literature classification. Table~\ref{tab:comparison} reports the comparative results.

\begin{table*}[t]
  \caption{Comparative experiment of AFPGNN method on Cora and Citeseer data sets.}
  \label{tab:comparison}
  \centering
  \small
  \begin{tabularx}{\textwidth}{@{}lYYYYYY@{}}
    \toprule
    & \multicolumn{3}{c}{Cora} & \multicolumn{3}{c}{Citeseer} \\
    \cmidrule(lr){2-4}\cmidrule(l){5-7}
    Method & Accuracy & Macro-F1 & Recall & Accuracy & Macro-F1 & Recall \\
    \midrule
    DeepWalk & 0.673 & 0.652 & 0.667 & 0.443 & 0.416 & 0.423 \\
    MLP & 0.561 & 0.515 & 0.503 & 0.465 & 0.455 & 0.451 \\
    GAE & 0.783 & 0.756 & 0.758 & 0.578 & 0.569 & 0.562 \\
    VGAE & 0.775 & 0.767 & 0.764 & 0.564 & 0.557 & 0.533 \\
    GCN & 0.815 & 0.781 & 0.788 & 0.703 & 0.686 & 0.632 \\
    MoNet & 0.811 & 0.779 & 0.787 & 0.691 & 0.672 & 0.611 \\
    GraphSAGE & 0.819 & 0.780 & 0.786 & 0.713 & 0.695 & 0.655 \\
    AFPGNN & 0.834 & 0.812 & 0.803 & 0.721 & 0.713 & 0.694 \\
    \bottomrule
  \end{tabularx}
\end{table*}

The comparison shows that AFPGNN is effective. On Cora and Citeseer, AFPGNN performs better than GCN, indicating that distinguishing weights among nodes in the same neighborhood is useful. The main reason is that the weighted sum of node features makes the weight of features in citation-related scientific documents depend on node features and document structure, whereas GCN is limited to two adjacent levels.

\subsection{Experiment 2: Ablation Experiment}

AFPGNN consists of three parts: adaptive feature processing, an encoder based on graph attention, and semantic learning based on maximum mutual information. AFPGNN-adaptive uses only adaptive feature processing for text classification. AFPGNN-attention uses only the graph attention encoder to learn representations. AFPGNN-mutual information maximizes mutual information between positive and negative samples based on contrastive learning. Hybrid variants combine two of the three components. Table~\ref{tab:ablation} shows the ablation results.

\begin{table*}[t]
  \caption{Results of ablation experiment.}
  \label{tab:ablation}
  \centering
  \small
  \begin{tabularx}{\textwidth}{@{}>{\raggedright\arraybackslash}p{0.30\textwidth}YYYYYY@{}}
    \toprule
    & \multicolumn{3}{c}{Cora} & \multicolumn{3}{c}{Citeseer} \\
    \cmidrule(lr){2-4}\cmidrule(l){5-7}
    Method & Accuracy & Macro-F1 & Recall & Accuracy & Macro-F1 & Recall \\
    \midrule
    AFPGNN-adaptive & 0.502 & 0.458 & 0.443 & 0.423 & 0.412 & 0.409 \\
    AFPGNN-attention & 0.825 & 0.789 & 0.786 & 0.671 & 0.653 & 0.621 \\
    AFPGNN-mutual information & 0.731 & 0.687 & 0.653 & 0.462 & 0.398 & 0.387 \\
    AFPGNN-adaptive+attention & 0.827 & 0.793 & 0.788 & 0.674 & 0.657 & 0.623 \\
    AFPGNN-adaptive+mutual information & 0.764 & 0.712 & 0.703 & 0.475 & 0.399 & 0.391 \\
    AFPGNN-attention+mutual information & 0.831 & 0.801 & 0.799 & 0.719 & 0.708 & 0.687 \\
    AFPGNN & 0.834 & 0.812 & 0.803 & 0.721 & 0.713 & 0.694 \\
    \bottomrule
  \end{tabularx}
\end{table*}

Table~\ref{tab:ablation} shows that AFPGNN performs better than the single-component variants, indicating that all three components are effective. By introducing graph attention, relationships among scientific-document node features are better integrated. The node representation is related only to adjacent nodes, which can be directly applied to inductive learning without obtaining the whole graph. Mutual information learning further improves the ability to learn scientific-document semantic representations.

\subsection{Experiment 3: Effect of Parameters}

The performance under different hyperparameters, such as alpha, dropout, learning rate, and the number of attention heads, is compared. The graph attention encoder is composed of $K=6$ attention heads, and each attention head calculates 10 features. The PReLU activation function adaptively learns parameters of the modified linear unit, and the alpha parameter is set to 0.2. In addition, dropout with $p=0.8$ is applied to the input of the graph attention encoder.

\begin{figure}[H]
  \centering
  \includegraphics[width=\linewidth]{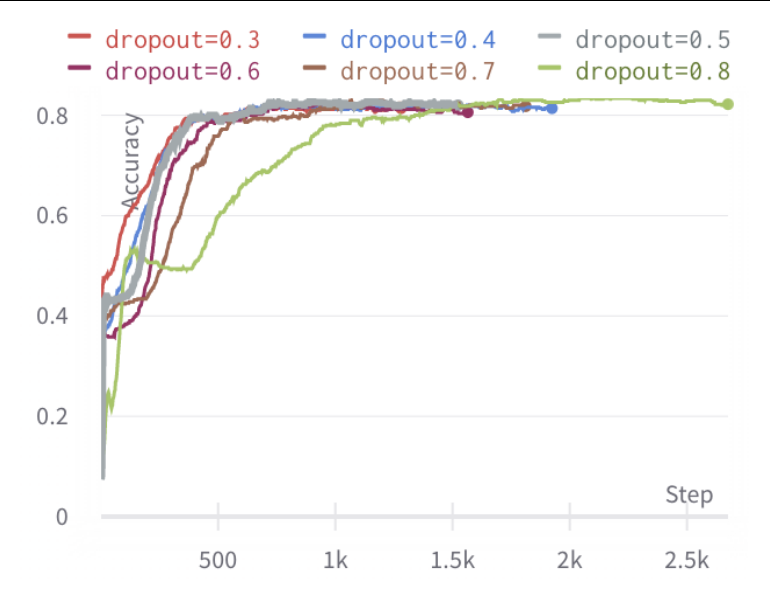}
  \Description{Line chart comparing dropout values from 0.3 to 0.8 over training steps.}
  \caption{The impact of dropout on semantic representation performance.}
  \label{fig:dropout}
\end{figure}

Figure~\ref{fig:dropout} shows the impact of dropout on model performance. In each training batch, overfitting can be reduced by ignoring neurons with a certain probability. When dropout equals 0.8, semantic representation learning performance is best and the training process is smoothest.

\begin{figure}[H]
  \centering
  \includegraphics[width=\linewidth]{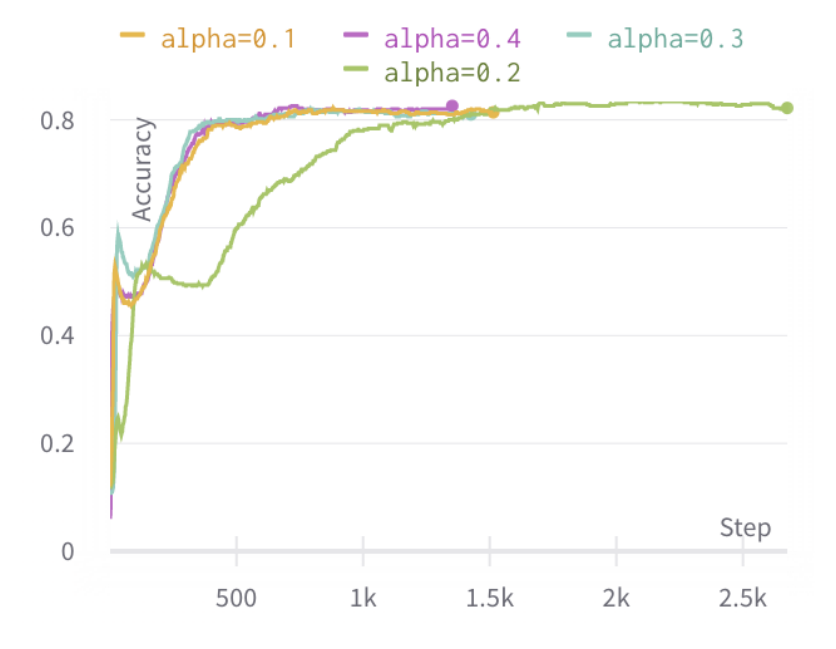}
  \Description{Line chart comparing alpha values over training steps.}
  \caption{The impact of alpha on semantic representation performance.}
  \label{fig:alpha}
\end{figure}

The influence of the attention coefficient alpha in Equation~(5) is shown in Figure~\ref{fig:alpha}. The PReLU activation function summarizes the traditional rectifier unit with negative slope, making the method more discriminative. When the same alpha is used for different channels, there are fewer parameters. PReLU treats alpha as a training parameter rather than a fixed slope. In this experiment, when alpha equals 0.2, the semantic representation learning effect is best and the training process is smoothest.

The effect of the number of attention heads is shown in Table~\ref{tab:heads}. Multi-head attention captures different attention information and integrates it to improve expressiveness. However, the number of attention heads is not necessarily better when larger, because more heads can increase cumulative error and reduce model performance. When the number of heads is 6, performance is best.

\begin{table}[t]
  \caption{The effect of the number of attention heads on the performance of semantic representation.}
  \label{tab:heads}
  \centering
  \small
  \begin{tabularx}{\linewidth}{@{}YYYY@{}}
    \toprule
    \multicolumn{4}{c}{Cora} \\
    \midrule
    Attention heads & Accuracy & Macro-F1 & Recall \\
    \midrule
    head=3 & 0.812 & 0.803 & 0.795 \\
    head=8 & 0.825 & 0.806 & 0.798 \\
    head=10 & 0.831 & 0.811 & 0.801 \\
    head=16 & 0.830 & 0.807 & 0.799 \\
    head=20 & 0.795 & 0.722 & 0.716 \\
    \textbf{head=6} & \textbf{0.834} & \textbf{0.812} & \textbf{0.803} \\
    \bottomrule
  \end{tabularx}
\end{table}

Negative sample features are generated by keeping the adjacency matrix unchanged and shuffling the features. The readout function converts local representation to graph-level implicit representation, and the discriminator maximizes local mutual information. Binary cross entropy is used as the loss. By reducing the loss, the discriminator correctly classifies scientific documents and increases JS divergence. The Adam SGD optimizer reduces cross entropy on training nodes, and the learning rate is set to 0.0001.

\begin{figure}[H]
  \centering
  \includegraphics[width=\linewidth]{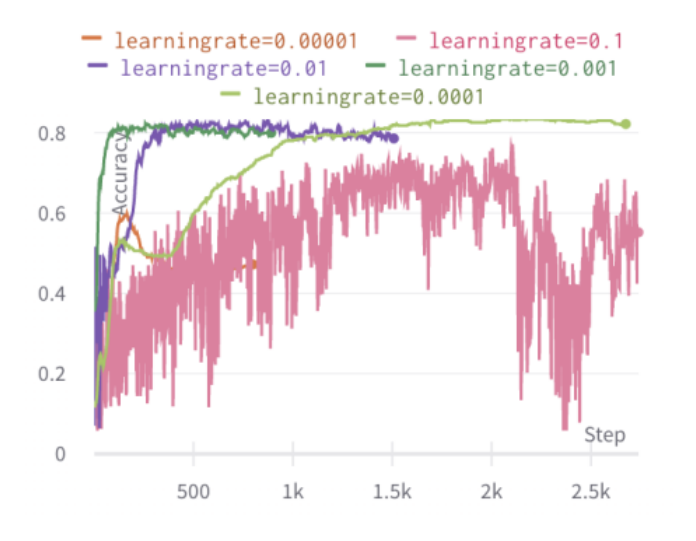}
  \Description{Line chart comparing learning rates over training steps.}
  \caption{The impact of learning rate on semantic representation performance.}
  \label{fig:learningrate}
\end{figure}

Figure~\ref{fig:learningrate} shows the effect of learning rate on semantic representation learning. The influence of output error on parameters increases with learning speed, and parameters update faster. When the learning rate is 0.0001, the semantic representation learning effect is best, the training process is relatively stable, and fluctuation is small.

\section{Conclusion}

This paper proposes an unsupervised learning method for semantic representation of scientific documents based on adaptive features and graph neural networks (AFPGNN). By using adaptive features, global and local information of scientific and technological documents are considered to enrich document features. By introducing graph attention, nodes are given different weights, and features of adjacent nodes are weighted and summed. An unsupervised graph neural network semantic representation learning method solves unlabeled and scalable large-graph problems. By comparing mutual information between positive and negative local semantic representations and the global graph semantic representation in latent space, the graph neural network captures local and global information, improving semantic representation learning. Experimental results show that AFPGNN achieves competitive performance in scientific literature classification.

\printbibliography

\end{document}